\documentclass{article} 
\usepackage{iclr2019_conference,times}

\usepackage[pdftex]{graphicx}
\usepackage{booktabs}
\usepackage{multirow}
\usepackage{amsmath}
\usepackage{bm}
\usepackage{stix}
\usepackage{relsize}
\usepackage{caption}
\usepackage{todonotes}
\usepackage{appendix}
\usepackage{hyperref}
\usepackage{url}

\def\eg{\emph{e.g.}}   
\def\ie{\emph{i.e.}}   
\def\cf{\emph{cf.}}     

\title{Fast Graph Representation Learning with\\PyTorch Geometric}

\author{Matthias Fey \& Jan E.~Lenssen\\
Department of Computer Graphics\\
TU Dortmund University\\
44227 Dortmund, Germany\\
\texttt{\{matthias.fey,janeric.lenssen\}@udo.edu}
}

\newcommand{\mr}[2]{\multirow{#1}{*}{#2}}
\newcommand{\rb}[2]{\rotatebox{#1}{#2}}

\iclrfinalcopy{}  
\begin{document}

\maketitle

\begin{abstract}
  We introduce \emph{PyTorch Geometric}, a library for deep learning on irregularly structured input data such as graphs, point clouds and manifolds, built upon PyTorch. In addition to general graph data structures and processing methods, it contains a variety of recently published methods from the domains of relational learning and 3D data processing.
  PyTorch Geometric achieves high data throughput by leveraging sparse GPU acceleration, by providing dedicated CUDA kernels and by introducing efficient mini-batch handling for input examples of different size.
  In this work, we present the library in detail and perform a comprehensive comparative study of the implemented methods in homogeneous evaluation scenarios.
\end{abstract}

\section{Introduction}

\emph{Graph Neural Networks} (GNNs) recently emerged as a powerful approach for representation learning on graphs, point clouds and manifolds \citep{Bronstein/etal/2017,Kipf/Welling/2017}.
Similar to the concepts of convolutional and pooling layers on regular domains, GNNs are able to (hierarchically) extract localized embeddings by passing, transforming, and aggregating information \citep{Bronstein/etal/2017,Gilmer/etal/2017,Battaglia/etal/2018,Ying/etal/2018,Morris/etal/2019}.

However, implementing GNNs is challenging, as high GPU throughput needs to be achieved on highly sparse and irregular data of varying size.
Here, we introduce \emph{PyTorch Geometric} (PyG), a geometric deep learning extension library for PyTorch \citep{Paszke/etal/2017} which achieves high performance by leveraging dedicated CUDA kernels. 
Following a simple message passing API, it bundles most of the recently proposed convolutional and pooling layers into a single and unified framework.
All implemented methods support both CPU and GPU computations and follow an immutable data flow paradigm that enables dynamic changes in graph structures through time.
PyG is released under the MIT license and is available on GitHub.\footnote{GitHub repository: \url{https://github.com/rusty1s/pytorch\_geometric}}
It is thoroughly documented and provides accompanying tutorials and examples as a first starting point.\footnote{Documentation: \url{https://rusty1s.github.io/pytorch\_geometric}}

\section{Overview}

In PyG, we represent a graph $\mathcal{G} = (\bm{X}, (\bm{I}, \bm{E}))$ by a node feature matrix $\bm{X} \in \mathbb{R}^{N \times F}$ of $N$ nodes and a sparse adjacency tuple $(\bm{I}, \bm{E})$ of $E$ edges, where $\bm{I} \in \mathbb{N}^{2 \times E}$ encodes edge indices in COOrdinate (COO) format and $\bm{E} \in \mathbb{R}^{E \times D}$ (optionally) holds $D$-dimensional edge features.
All user facing APIs, \eg, data loading routines, multi-GPU support, data augmentation or model instantiations are heavily inspired by PyTorch to keep them as familiar as possible.

\paragraph{Neighborhood Aggregation.}

Generalizing the convolutional operator to irregular domains is typically expressed as a \emph{neighborhood aggregation} or \emph{message passing} scheme \citep{Gilmer/etal/2017}
\begin{equation}
  \vec{x}_i^{\prime} = \gamma \left( \vec{x}_i, \operatorname*{\mathlarger{\dottedsquare}}\limits_{j \in \mathcal{N}(i)} \phi \left(\vec{x}_i, \,\, \vec{x}_j, \,\, \vec{e}_{j,i} \right) \right)
\end{equation}
where $\dottedsquare$ denotes a differentiable, permutation invariant function, \eg, sum, mean or max, and $\gamma$ and $\phi$ denote differentiable functions, \eg, MLPs.
\begin{figure}[t]
  \centering
  \includegraphics[width=\linewidth]{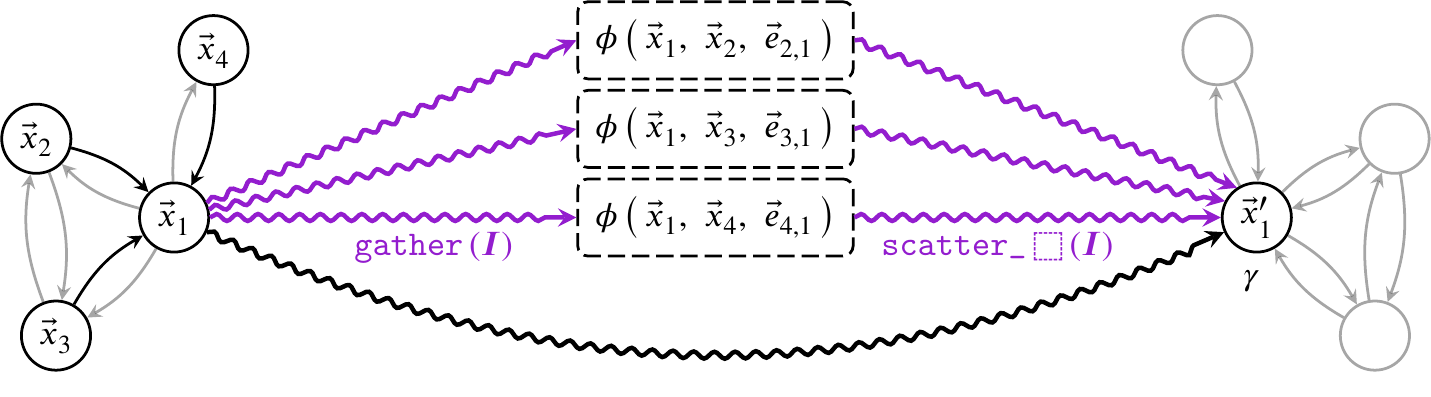}
  \caption{Computation scheme of a GNN layer by leveraging gather and scatter methods based on edge indices $\bm{I}$, hence alternating between node parallel space and edge parallel space.}\label{fig:gather_scatter}
\end{figure}
In practice, this can be achieved by gathering and scattering of node features and vectorized element-wise computation of $\gamma$ and $\phi$, as visualized in Figure~\ref{fig:gather_scatter}.
Although working on irregularly structured input, this scheme can be heavily accelerated by the GPU\@.
In contrast to implementations via sparse matrix multiplications, the usage of gather/scatter proves to be advantageous for low-degree graphs and non-coalesced input (\cf~Appendix~\ref{sec:gather_and_scatter}), and allows for the integration of central node and multi-dimensional edge information while aggregating.

We provide the user with a general \texttt{MessagePassing} interface to allow for rapid and clean prototyping of new research ideas.
In order to use, users only need to define the methods $\phi$, \ie, \texttt{message}, and $\gamma$, \ie, \texttt{update}, as well as chosing an aggregation scheme $\dottedsquare$.
For implementing $\phi$, node features are automatically mapped to the respective source and target nodes.

Almost all recently proposed neighborhood aggregation functions can be lifted to this interface, including (but not limited to) the methods already integrated into PyG\@:
For learning on arbitrary graphs we have implemented GCN \citep{Kipf/Welling/2017} and its simplified version (SGC) from \citet{Wu/etal/2019},
the spectral chebyshev and ARMA filter convolutions \citep{Defferrard/etal/2016, Bianchi/etal/2019},
GraphSAGE \citep{Hamilton/etal/2017},
the attention-based operators GAT \citep{Velickovic/etal/2018} and AGNN \citep{Thekumparampil/etal/2018},
the Graph Isomorphism Network (GIN) from \citet{Xu/etal/2019},
the Approximate Personalized Propagation of Neural Predictions (APPNP) operator \citep{Klicpera/etal/2019},
the Dynamic Neighborhood Aggregation (DNA) operator \citep{Fey/2019}
and the signed operator for learning in signed networks \citep{Derr/etal/2018}.

For learning on point clouds, manifolds and graphs with multi-dimensional edge features, we provide
the relational GCN operator from \citet{Schlichtkrull/etal/2018},
PointNet++ \citep{Qi/etal/2017b},
PointCNN \citep{Li/etal/2018},
and the continuous kernel-based methods
MPNN \citep{Gilmer/etal/2017,Simonovsky/Komodakis/2017},
MoNet \citep{Monti/etal/2017},
SplineCNN \citep{Fey/etal/2018} and
the edge convolution operator (EdgeCNN) from \citet{Wang/etal/2018a}.

In addition to these operators, we provide high-level implementations of, \eg, maximizing mutual information \citep{Velickovic/etal/2019}, autoencoding graphs \citep{Kipf/Welling/2016, Pan/etal/2018}, aggregating jumping knowledge \citep{Xu/etal/2018}, and predicting temporal events in knowledge graphs \citep{Jin/etal/2019}.

\paragraph{Global Pooling.}

PyG also supports graph-level outputs as opposed to node-level outputs by providing a variety of \emph{readout} functions such as global add, mean or max pooling.
We additionaly offer more sophisticated methods such as set-to-set \citep{Vinyals/etal/2016}, sort pooling \citep{Zhang/etal/2018} or the global soft attention layer from \citet{Li/etal/2016}.

\paragraph{Hierarchical Pooling.}

To further extract hierarchical information and to allow deeper GNN models, various pooling approaches can be applied in a spatial or data-dependent manner.
We currently provide implementation examples for Graclus \citep{Dhillon/etal/2007,Fagginger/etal/2011} and
voxel grid pooling \citep{Simonovsky/Komodakis/2017},
the iterative farthest point sampling algorithm \citep{Qi/etal/2017b} followed by $k$-NN or query ball graph generation \citep{Qi/etal/2017b,Wang/etal/2018a},
and differentiable pooling mechanisms such as DiffPool \citep{Ying/etal/2018}
and $\mathrm{top}_k$ pooling \citep{Gao/Ji/2018,Cangea/etal/2018}.

\paragraph{Mini-batch Handling.}

Our framework supports batches of multiple graph instances (of potentially different size) by automatically creating a single (sparse) block-diagonal adjacency matrix and concatenating feature matrices in the node dimension.
Therefore, neighborhood aggregation methods can be applied without modification, since no messages are exchanged between disconnected graphs.
In addition, an automatically generated assignment vector ensures that node-level information is not aggregated across graphs, \eg, when executing global aggregation operators.

\paragraph{Processing of Datasets.}

We provide a consistent data format and an easy-to-use interface for the creation and processing of datasets, both for large datasets and for datasets that can be kept in memory during training.
In order to create new datasets, users just need to read/download their data and convert it to the PyG data format in the respective \texttt{process} method.
In addition, datasets can be modified by the use of \texttt{transforms}, which take in separate graphs and transform them, \eg, for data augmentation, for enhancing node features with synthetic structural graph properties \citep{Cai/Wang/2018}, to automatically generate graphs from point clouds or to sample point clouds from meshes.

PyG already supports a lot of common benchmark datasets often found in literature which are automatically downloaded and processed on first instantiation. In detail, we provide over 60 graph kernel benchmark datasets\footnote{Kernel datasets: \url{http://graphkernels.cs.tu-dortmund.de}} \citep{Kersting/etal/2016}, \eg, PROTEINS or IMDB-BINARY,
the citation graphs Cora, CiteSeer, PubMed and Cora-Full \citep{Sen/etal/2008,Bojchevski/Guennemann/2018},
the Coauthor CS/Physics and Amazon Computers/Photo datasets from \citet{Shchur/etal/2018},
the molecule datasets QM7b \citep{Montavon/etal/2013} and QM9 \citep{Ramakrishnan/etal/2014},
the protein-protein interaction graphs from \citet{Hamilton/etal/2017}, and
the temporal datasets Bitcoin-OTC \citep{Kumar/etal/2016}, ICEWS \citep{Boschee/etal/2015} and GDELT \citep{Leetaru/Schrodt/2013}.
In addition, we provide embedded datasets like MNIST superpixels \citep{Monti/etal/2017},
FAUST \citep{Bogo/etal/2014},
ModelNet10/40 \citep{Wu/etal/2015},
ShapeNet \citep{Chang/etal/2015},
COMA \citep{Ranjan/etal/2018} and
the PCPNet dataset from \citet{Guerrero/etal/2018}.

\section{Empirical Evaluation}

We evaluate the correctness of the implemented methods by performing a comprehensive comparative study in homogeneous evaluation scenarios.
Descriptions and statistics of all used datasets can be found in Appendix~\ref{sec:datasets}.
For all experiments, we tried to follow the hyperparameter setup of the respective papers as closely as possible.
The individual experimental setups can be derived and all experiments can be replicated from the code provided at our GitHub repository.\footnote{\url{https://github.com/rusty1s/pytorch\_geometric/benchmark}}

\paragraph{Semi-supervised Node Classification.}

\begin{table}[t]
  \centering
  \caption{Semi-supervised node classification with both fixed and random splits.}\label{tab:node_classification}
  \begin{tabular}{lcccccc}
    \toprule
      \mr{2}{\textbf{Method}} & \multicolumn{2}{c}{\textbf{Cora}} & \multicolumn{2}{c}{\textbf{CiteSeer}} & \multicolumn{2}{c}{\textbf{PubMed}}\\
      & Fixed & Random & Fixed & Random & Fixed & Random\\
    \midrule
      Cheby & 81.4 $\pm$ 0.7          & 77.8 $\pm$ 2.2          & 70.2 $\pm$ 1.0          & 67.7 $\pm$ 1.7          & 78.4 $\pm$ 0.4          & 75.8 $\pm$ 2.2          \\ 
      GCN   & 81.5 $\pm$ 0.6          & 79.4 $\pm$ 1.9          & 71.1 $\pm$ 0.7          & 68.1 $\pm$ 1.7          & 79.0 $\pm$ 0.6          & 77.4 $\pm$ 2.4          \\ 
      GAT   & 83.1 $\pm$ 0.4          & 81.0 $\pm$ 1.4          & 70.8 $\pm$ 0.5          & 69.2 $\pm$ 1.9          & 78.5 $\pm$ 0.3          & 78.3 $\pm$ 2.3          \\ 
      SGC   & 81.7 $\pm$ 0.1          & 80.2 $\pm$ 1.6          & 71.3 $\pm$ 0.2          & 68.7 $\pm$ 1.6          & 78.9 $\pm$ 0.1          & 76.5 $\pm$ 2.4          \\ 
      ARMA  & 82.8 $\pm$ 0.6          & 80.7 $\pm$ 1.4          & \textbf{72.3 $\pm$ 1.1} & 68.9 $\pm$ 1.6          & 78.8 $\pm$ 0.3          & 77.7 $\pm$ 2.6          \\ 
      APPNP & \textbf{83.3 $\pm$ 0.5} & \textbf{82.2 $\pm$ 1.5} & 71.8 $\pm$ 0.5          & \textbf{70.0 $\pm$ 1.4} & \textbf{80.1 $\pm$ 0.2} & \textbf{79.4 $\pm$ 2.2} \\ 
    \bottomrule
  \end{tabular}
\end{table}

We perform semi-supervised node classification (\cf~Table~\ref{tab:node_classification}) by reporting average accuracies of (a) 100 runs for the fixed train/val/test split from \citet{Kipf/Welling/2017}, and (b) 100 runs of randomly initialized train/val/test splits as suggested by \citet{Shchur/etal/2018}, where we additionally ensure uniform class distribution on the train split.

Nearly all experiments show a high reproducibility of the results reported in the respective papers.
However, test performance is worse for all models when using random data splits.
Among the experiments, the APPNP operator \citep{Klicpera/etal/2019} generally performs best, while the ARMA \citep{Bianchi/etal/2019}, SGC \citep{Wu/etal/2019}, GCN \citep{Kipf/Welling/2017} and GAT \citep{Velickovic/etal/2018} operators follow closely behind.

\paragraph{Graph Classification.}

\begin{table}[t]
  \centering
  \caption{Graph classification.}\label{tab:graph_classification}
  \begin{tabular}{llccccc}
    \toprule
      & \mr{2}{\textbf{Method}} & \mr{2}{\textbf{MUTAG}} & \mr{2}{\textbf{PROTEINS}} & \mr{2}{\textbf{COLLAB}} & \textbf{IMDB-} & \textbf{REDDIT-}\\
      & & & & & \textbf{BINARY} & \textbf{BINARY}\\
    \midrule
      \mr{4}{\rb{90}{Flat}} & GCN            & 74.6 $\pm$ 7.7          & 73.1 $\pm$ 3.8          & \textbf{80.6 $\pm$ 2.1} & 72.6 $\pm$ 4.5          & 89.3 $\pm$ 3.3 \\ 
      & SAGE           & 74.9 $\pm$ 8.7          & \textbf{73.8 $\pm$ 3.6} & 79.7 $\pm$ 1.7          & 72.4 $\pm$ 3.6          & 89.1 $\pm$ 1.9 \\ 
      & GIN-0          & \textbf{85.7 $\pm$ 7.7}          & 72.1 $\pm$ 5.1          & 79.3 $\pm$ 2.7          & \textbf{72.8 $\pm$ 4.5} & 89.6 $\pm$ 2.6 \\ 
      & GIN-$\epsilon$ & 83.4 $\pm$ 7.5 & 72.6 $\pm$ 4.9          & 79.8 $\pm$ 2.4          & 72.1 $\pm$ 5.1          & \textbf{90.3 $\pm$ 3.0} \\ 
    \midrule
      \mr{3}{\rb{90}{Hier.}} & Graclus          & 77.1 $\pm$ 7.2 & 73.0 $\pm$ 4.1 & 79.6 $\pm$ 2.0 & 72.2 $\pm$ 4.2 & 88.8 $\pm$ 3.2 \\ 
      & $\mathrm{top}_k$ & 76.3 $\pm$ 7.5  & 72.7 $\pm$ 4.1 & \textbf{79.7 $\pm$ 2.2} & 72.5 $\pm$ 4.6 & 87.6 $\pm$ 2.4 \\ 
      & DiffPool         & \textbf{85.0 $\pm$ 10.3} & \textbf{75.1 $\pm$ 3.5} & 78.9 $\pm$ 2.3 & \textbf{72.6 $\pm$ 3.9} & \textbf{92.1 $\pm$ 2.6} \\ 
    \midrule
      \mr{4}{\rb{90}{Global}} & SAGE w/o JK     & 73.7 $\pm$ 7.8          & 72.7 $\pm$ 3.6          & 79.6 $\pm$ 2.4          & 72.1 $\pm$ 4.4          & 87.9 $\pm$ 1.9 \\ 
      & GlobalAttention & 74.6 $\pm$ 8.0          & 72.5 $\pm$ 4.5          & \textbf{79.6 $\pm$ 2.2} & 72.3 $\pm$ 3.8          & 87.4 $\pm$ 2.5 \\ 
      & Set2Set         & 73.7 $\pm$ 6.9          & \textbf{73.6 $\pm$ 3.7} & 79.6 $\pm$ 2.3          & 72.2 $\pm$ 4.2          & \textbf{89.6 $\pm$ 2.4} \\ 
      & SortPool        & \textbf{77.3 $\pm$ 8.9} & 72.4 $\pm$ 4.1          & 77.7 $\pm$ 3.1          & \textbf{72.4 $\pm$ 3.8} & 74.9 $\pm$ 6.7 \\ 
    \bottomrule
  \end{tabular}
\end{table}

We report the average accuracy of 10-fold cross validation on a number of common benchmark datasets (\cf~Table~\ref{tab:graph_classification}) where we randomly sample a training fold to serve as a validation set.
We only make use of discrete node features.
In case they are not given, we use one-hot encodings of node degrees as feature input.
For all experiments, we use the global mean operator to obtain graph-level outputs.
Inspired by the Jumping Knowledge framework \citep{Xu/etal/2018}, we compute graph-level outputs after each convolutional layer and combine them via concatenation.
For evaluating the (global) pooling operators, we use the GraphSAGE operator as our baseline.
We omit Jumping Knowledge when comparing global pooling operators, and hence report an additional baseline based on global mean pooling.
For each dataset, we tune (1) the number of hidden units $\in \{ 16, 32, 64, 128 \}$ and (2) the number of layers $\in \{ 2, 3, 4, 5 \}$ with respect to the validation set.

Due to standardized evaluations and network architectures, not all results are aligned with their official reported values.
For example, except for DiffPool \citep{Ying/etal/2018}, (global) pooling operators do not perform as benefically as expected to their respective (flat) counterparts, especially when baselines are enhanced by Jumping Knowledge \citep{Xu/etal/2018}.
However, the potential of more sophisticated approaches may not be well-reflected on these simple benchmark tasks \citep{Cai/Wang/2018}.
Among the flat GNN approaches, the GIN layer \citep{Xu/etal/2019} generally achieves the best results.

\begin{table}[t]
  \centering
  \begin{minipage}[c]{0.35\linewidth}
    \centering
    \caption{Point cloud classification.}\label{tab:point_cloud_classification}
    \begin{tabular}{lc}
      \toprule
        \textbf{Method} & \textbf{ModelNet10}\\
      \midrule
        MPNN & 92.07 \\ 
        PointNet++ & 92.51 \\ 
        EdgeCNN & 92.62 \\ 
        SplineCNN & 92.65 \\ 
        PointCNN & \textbf{93.28} \\ 
      \bottomrule
    \end{tabular}
  \end{minipage}
  \hspace{0.5cm}
  \begin{minipage}[c]{0.52\linewidth}
    \centering
    \caption{Training runtime comparison.}\label{tab:training_runtimes}
    \begin{tabular}{lcrrr}
      \toprule
        \mr{2}{\textbf{Dataset}} & \mr{2}{\textbf{Method}} & \textbf{DGL} & \textbf{DGL} & \mr{2}{\textbf{PyG}} \\
        & & \textbf{DB~\,} & \textbf{GS~\,} & \\
      \midrule
        \mr{2}{Cora}     & GCN  & 4.19s  & 0.32s  & \textbf{0.25s} \\ 
                         & GAT  & 6.31s  & 5.36s  & \textbf{0.80s} \\ 
      \midrule
        \mr{2}{CiteSeer} & GCN  & 3.78s  & 0.34s  & \textbf{0.30s} \\ 
                         & GAT  & 5.61s  & 4.91s  & \textbf{0.88s} \\ 
      \midrule
        \mr{2}{PubMed}   & GCN  & 12.91s & 0.36s  & \textbf{0.32s} \\ 
                         & GAT  & 18.69s & 13.76s & \textbf{2.42s} \\ 
      \midrule
        MUTAG            & RGCN & 18.81s & 2.40s  & \textbf{2.14s} \\ 
      \bottomrule
    \end{tabular}
  \end{minipage}
\end{table}

\paragraph{Point Cloud Classification.}

We evaluate various point cloud methods on ModelNet10 \citep{Wu/etal/2015} where we uniformly sample 1,024 points from mesh surfaces based on face area (\cf~Table~\ref{tab:point_cloud_classification}).
As hierarchical pooling layers, we use the iterative farthest point sampling algorithm followed by a new graph generation based on a larger query ball (PointNet++ \citep{Qi/etal/2017b}, MPNN \citep{Gilmer/etal/2017,Simonovsky/Komodakis/2017} and SplineCNN \citep{Fey/etal/2018}) or based on a fixed number of nearest neighbors (EdgeCNN \citep{Wang/etal/2018a} and PointCNN \citep{Li/etal/2018}).
We have taken care to use approximately the same number of parameters for each model.

All approaches perform nearly identically with PointCNN~\citep{Li/etal/2018} taking a slight lead.
We attribute this to the fact that all operators are based on similar principles and might have the same expressive power for the given task.

\paragraph{Runtime Experiments.}

We conduct several experiments on a number of dataset-model pairs to report the runtime of a whole training procedure for 200 epochs obtained on a single NVIDIA GTX 1080 Ti (\cf~Table~\ref{tab:training_runtimes}).
As it shows, PyG is very fast despite working on sparse data.
Compared to the \emph{Degree Bucketing} (DB) approach of the \emph{Deep Graph Library} (DGL) v0.2 \citep{Wang/etal/2018b}, PyG trains models up to 40 times faster.
Although runtimes are comparable when using gather and scatter optimizations (GS) inside DGL, we could further improve runtimes of GAT \citep{Velickovic/etal/2018} by up to 7 times by providing our own optimized sparse softmax kernels.

\section{Roadmap and Conclusion}

We presented the PyTorch Geometric framework for fast representation learning on graphs, point clouds and manifolds.
We are actively working to further integrate existing methods and plan to quickly integrate future methods into our framework.
All researchers and software engineers are invited to collaborate with us in extending its scope.

\subsubsection*{Acknowledgments}

This work has been supported by the \emph{German Research Association (DFG)} within the Collaborative Research Center SFB 876, \emph{Providing Information by Resource-Constrained Analysis}, projects A6 and B2.
We thank Moritz Ludolph and all other contributors for their amazing involvement in this project.
Last but not least, we thank Christopher Morris for fruitful discussions, proofreading and helpful advice.

\bibliography{iclr2019_conference}

\begin{thebibliography}{52}
\providecommand{\natexlab}[1]{#1}
\providecommand{\url}[1]{\texttt{#1}}
\expandafter\ifx\csname urlstyle\endcsname\relax
  \providecommand{\doi}[1]{doi: #1}\else
  \providecommand{\doi}{doi: \begingroup \urlstyle{rm}\Url}\fi

\bibitem[Battaglia et~al.(2018)Battaglia, Hamrick, Bapst, Sanchez{-}Gonzalez,
  Zambaldi, Malinowski, Tacchetti, Raposo, Santoro, Faulkner,
  G{\"{u}}l{\c{c}}ehre, Song, Ballard, Gilmer, Dahl, Vaswani, Allen, Nash,
  Langston, Dyer, Heess, Wierstra, Kohli, Botvinick, Vinyals, Li, and
  Pascanu]{Battaglia/etal/2018}
P.~W. Battaglia, J.~B. Hamrick, V.~Bapst, A.~Sanchez{-}Gonzalez, V.~F.
  Zambaldi, M.~Malinowski, A.~Tacchetti, D.~Raposo, A.~Santoro, R.~Faulkner,
  {\c{C}}.~G{\"{u}}l{\c{c}}ehre, F.~Song, A.~J. Ballard, J.~Gilmer, G.~E. Dahl,
  A.~Vaswani, K.~Allen, C.~Nash, V.~Langston, C.~Dyer, N.~Heess, D.~Wierstra,
  P.~Kohli, M.~Botvinick, O.~Vinyals, Y.~Li, and R.~Pascanu.
\newblock Relational inductive biases, deep learning, and graph networks.
\newblock \emph{CoRR}, abs/1806.01261, 2018.

\bibitem[Bianchi et~al.(2019)Bianchi, Grattarola, Livi, and
  Alippi]{Bianchi/etal/2019}
F.~M. Bianchi, D.~Grattarola, L.~Livi, and C.~Alippi.
\newblock Graph neural networks with convolutional {ARMA} filters.
\newblock \emph{CoRR}, abs/1901.01343, 2019.

\bibitem[Bogo et~al.(2014)Bogo, Romero, Loper, and Black]{Bogo/etal/2014}
F.~Bogo, J.~Romero, M.~Loper, and M.~J. Black.
\newblock {FAUST}: Dataset and evaluation for {3D} mesh registration.
\newblock In \emph{CVPR}, 2014.

\bibitem[Bojchevski \& G{\"u}nnemann(2018)Bojchevski and
  G{\"u}nnemann]{Bojchevski/Guennemann/2018}
A.~Bojchevski and S.~G{\"u}nnemann.
\newblock Deep gaussian embedding of attributed graphs: Unsupervised inductive
  learning via ranking.
\newblock In \emph{ICLR}, 2018.

\bibitem[Boschee et~al.(2015)Boschee, Lautenschlager, O'Brien, Shellman, Starz,
  and Ward]{Boschee/etal/2015}
E.~Boschee, J.~Lautenschlager, S.~O'Brien, S.~Shellman, J.~Starz, and M.~Ward.
\newblock {ICEWS} coded event data.
\newblock \emph{Harvard Dataverse}, 2015.

\bibitem[Bronstein et~al.(2017)Bronstein, Bruna, LeCun, Szlam, and
  Vandergheynst]{Bronstein/etal/2017}
M.~M. Bronstein, J.~Bruna, Y.~LeCun, A.~Szlam, and P.~Vandergheynst.
\newblock Geometric deep learning: Going beyond euclidean data.
\newblock In \emph{Signal Processing Magazine}, 2017.

\bibitem[Cai \& Wang(2018)Cai and Wang]{Cai/Wang/2018}
C.~Cai and Y.~Wang.
\newblock A simple yet effective baseline for non-attribute graph
  classification.
\newblock \emph{CoRR}, abs/1811.03508, 2018.

\bibitem[Cangea et~al.(2018)Cangea, Veli{\v{c}}kovi{\'{c}}, Jovanovi{\'{c}},
  Kipf, and Li{\`{o}}]{Cangea/etal/2018}
C.~Cangea, P.~Veli{\v{c}}kovi{\'{c}}, N.~Jovanovi{\'{c}}, T.~N. Kipf, and
  P.~Li{\`{o}}.
\newblock Towards sparse hierarchical graph classifiers.
\newblock In \emph{NeurIPS-W}, 2018.

\bibitem[Chang et~al.(2015)Chang, Funkhouser, Guibas, Hanrahan, Huang, Li,
  Savarese, Savva, Song, Su, Xiao, Yi, and Yu]{Chang/etal/2015}
A.~X. Chang, T.~Funkhouser, L.~J. Guibas, P.~Hanrahan, Q.~Huang, Z.~Li,
  S.~Savarese, M.~Savva, S.~Song, H.~Su, J.~Xiao, L.~Yi, and F.~Yu.
\newblock {ShapeNet}: An information-rich {3D} model repository.
\newblock \emph{CoRR}, abs/1512.03012, 2015.

\bibitem[Defferrard et~al.(2016)Defferrard, Bresson, and
  Vandergheynst]{Defferrard/etal/2016}
M.~Defferrard, X.~Bresson, and P.~Vandergheynst.
\newblock Convolutional neural networks on graphs with fast localized spectral
  filtering.
\newblock In \emph{NIPS}, 2016.

\bibitem[Derr et~al.(2018)Derr, Ma, and Tang]{Derr/etal/2018}
T.~Derr, Y.~Ma, and J.~Tang.
\newblock Signed graph convolutional networks.
\newblock In \emph{ICDM}, 2018.

\bibitem[Dhillon et~al.(2007)Dhillon, Guan, and Kulis]{Dhillon/etal/2007}
I.~S. Dhillon, Y.~Guan, and B.~Kulis.
\newblock Weighted graph cuts without eigenvectors: A multilevel approach.
\newblock In \emph{TPAMI}, 2007.

\bibitem[Fagginger~Auer \& Bisseling(2011)Fagginger~Auer and
  Bisseling]{Fagginger/etal/2011}
B.~O. Fagginger~Auer and R.~H. Bisseling.
\newblock A {GPU} algorithm for greedy graph matching.
\newblock In \emph{Facing the Multicore - Challenge {II} - Aspects of New
  Paradigms and Technologies in Parallel Computing}, 2011.

\bibitem[Fey(2019)]{Fey/2019}
M.~Fey.
\newblock Just jump: Dynamic neighborhood aggregation in graph neural networks.
\newblock In \emph{ICLR-W}, 2019.

\bibitem[Fey et~al.(2018)Fey, Lenssen, Weichert, and M{\"u}ller]{Fey/etal/2018}
M.~Fey, J.~E. Lenssen, F.~Weichert, and H.~M{\"u}ller.
\newblock {SplineCNN}: Fast geometric deep learning with continuous {B}-spline
  kernels.
\newblock In \emph{CVPR}, 2018.

\bibitem[Gao \& Ji(2018)Gao and Ji]{Gao/Ji/2018}
H.~Gao and S.~Ji.
\newblock Graph {U}-{N}et.
\newblock \url{https://openreview.net/forum?id=HJePRoAct7}, 2018.
\newblock Submitted to ICLR.

\bibitem[Gilmer et~al.(2017)Gilmer, Schoenholz, Riley, Vinyals, and
  Dahl]{Gilmer/etal/2017}
J.~Gilmer, S.~S. Schoenholz, P.~F. Riley, O.~Vinyals, and G.~E. Dahl.
\newblock Neural message passing for quantum chemistry.
\newblock In \emph{ICML}, 2017.

\bibitem[Guerrero et~al.(2018)Guerrero, Kleiman, Ovsjanikov, and
  Mitra]{Guerrero/etal/2018}
P.~Guerrero, Y.~Kleiman, M.~Ovsjanikov, and N.~J. Mitra.
\newblock {PCPNet}: Learning local shape properties from raw point clouds.
\newblock \emph{Computer Graphics Forum}, 37, 2018.

\bibitem[Hamilton et~al.(2017)Hamilton, Ying, and Leskovec]{Hamilton/etal/2017}
W.~L. Hamilton, R.~Ying, and J.~Leskovec.
\newblock Inductive representation learning on large graphs.
\newblock In \emph{NIPS}, 2017.

\bibitem[Jin et~al.(2019)Jin, Zhang, Szekely, and Ren]{Jin/etal/2019}
W.~Jin, C.~Zhang, P.~Szekely, and X.~Ren.
\newblock Recurrent event network for reasoning over temporal knowledge graphs.
\newblock In \emph{ICLR-W}, 2019.

\bibitem[Kersting et~al.(2016)Kersting, Kriege, Morris, Mutzel, and
  Neumann]{Kersting/etal/2016}
K.~Kersting, N.~M. Kriege, C.~Morris, P.~Mutzel, and M.~Neumann.
\newblock Benchmark data sets for graph kernels.
\newblock \url{http://graphkernels.cs.tu-dortmund.de}, 2016.

\bibitem[Kipf \& Welling(2016)Kipf and Welling]{Kipf/Welling/2016}
T.~N. Kipf and M.~Welling.
\newblock Variational graph auto-encoders.
\newblock In \emph{NIPS-W}, 2016.

\bibitem[Kipf \& Welling(2017)Kipf and Welling]{Kipf/Welling/2017}
T.~N. Kipf and M.~Welling.
\newblock Semi-supervised classification with graph convolutional networks.
\newblock In \emph{ICLR}, 2017.

\bibitem[Klicpera et~al.(2019)Klicpera, Bojchevski, and
  G{\"u}nnemann]{Klicpera/etal/2019}
J.~Klicpera, A.~Bojchevski, and S.~G{\"u}nnemann.
\newblock Predict then propagate: Graph neural networks meet personalized
  {PageRank}.
\newblock In \emph{ICLR}, 2019.

\bibitem[Kumar et~al.(2016)Kumar, Spezzano, Subrahmanian, and
  Faloutsos]{Kumar/etal/2016}
S.~Kumar, F.~Spezzano, V.~Subrahmanian, and C.~Faloutsos.
\newblock Edge weight prediction in weighted signed networks.
\newblock In \emph{ICDM}, 2016.

\bibitem[Leetaru \& Schrodt(2013)Leetaru and Schrodt]{Leetaru/Schrodt/2013}
K.~Leetaru and P.~A. Schrodt.
\newblock {GDELT}: Global data on events, location, and tone.
\newblock \emph{ISA Annual Convention}, 2013.

\bibitem[Li et~al.(2016)Li, Tarlow, Brockschmidt, and Zemel]{Li/etal/2016}
Y.~Li, D.~Tarlow, M.~Brockschmidt, and R.~Zemel.
\newblock Gated graph sequence neural networks.
\newblock In \emph{ICLR}, 2016.

\bibitem[Li et~al.(2018)Li, Bu, Sun, Wu, Di, and Chen]{Li/etal/2018}
Y.~Li, R.~Bu, M.~Sun, W.~Wu, X.~Di, and B.~Chen.
\newblock {PointCNN}: Convolution on {$\mathcal{X}$}-transformed points.
\newblock In \emph{NeurIPS}, 2018.

\bibitem[Montavon et~al.(2013)Montavon, Rupp, Gobre, Vazquez-Mayagoitia,
  Hansen, Tkatchenko, M{\"u}ller, and von Lilienfeld]{Montavon/etal/2013}
G.~Montavon, M.~Rupp, V.~Gobre, A.~Vazquez-Mayagoitia, K.~Hansen,
  A.~Tkatchenko, K.~M{\"u}ller, and O.~A. von Lilienfeld.
\newblock Machine learning of molecular electronic properties in chemical
  compound space.
\newblock \emph{New Journal of Physics}, 2013.

\bibitem[Monti et~al.(2017)Monti, Boscaini, Masci, Rodol{\`{a}}, Svoboda, and
  Bronstein]{Monti/etal/2017}
F.~Monti, D.~Boscaini, J.~Masci, E.~Rodol{\`{a}}, J.~Svoboda, and M.~M.
  Bronstein.
\newblock Geometric deep learning on graphs and manifolds using mixture model
  {CNN}s.
\newblock In \emph{CVPR}, 2017.

\bibitem[Morris et~al.(2019)Morris, Ritzert, Fey, Hamilton, Lenssen, Rattan,
  and Grohe]{Morris/etal/2019}
C.~Morris, M.~Ritzert, M.~Fey, W.~L. Hamilton, J.~E. Lenssen, G.~Rattan, and
  M.~Grohe.
\newblock {W}eisfeiler and {L}eman go neural: Higher-order graph neural
  networks.
\newblock In \emph{AAAI}, 2019.

\bibitem[Pan et~al.(2018)Pan, Hu, Long, Jiang, Yao, and Zhang]{Pan/etal/2018}
S.~Pan, R.~Hu, G.~Long, J.~Jiang, L.~Yao, and C.~Zhang.
\newblock Adversarially regularized graph autoencoder for graph embedding.
\newblock In \emph{IJCAI}, 2018.

\bibitem[Paszke et~al.(2017)Paszke, Gross, Chintala, Chanan, Yang, DeVito, Lin,
  Desmaison, Antiga, and Lerer]{Paszke/etal/2017}
A.~Paszke, S.~Gross, S.~Chintala, G.~Chanan, E.~Yang, Z.~DeVito, Z.~Lin,
  A.~Desmaison, L.~Antiga, and A.~Lerer.
\newblock Automatic differentiation in {P}y{T}orch.
\newblock In \emph{NIPS-W}, 2017.

\bibitem[Qi et~al.(2017)Qi, Yi, Su, and Guibas]{Qi/etal/2017b}
C.~R. Qi, L.~Yi, H.~Su, and L.~J. Guibas.
\newblock {PointNet++}: Deep hierarchical feature learning on point sets in a
  metric space.
\newblock In \emph{NIPS}, 2017.

\bibitem[Ramakrishnan et~al.(2014)Ramakrishnan, Dral, Rupp, and von
  Lilienfeld]{Ramakrishnan/etal/2014}
R.~Ramakrishnan, P.~O. Dral, M.~Rupp, and O.~A. von Lilienfeld.
\newblock Quantum chemistry structures and properties of 134 kilo molecules.
\newblock \emph{Scientific Data}, 2014.

\bibitem[Ranjan et~al.(2018)Ranjan, Bolkart, Sanyal, and
  Black]{Ranjan/etal/2018}
A.~Ranjan, T.~Bolkart, S.~Sanyal, and M.~J. Black.
\newblock Generating {3D} faces using convolutional mesh autoencoders.
\newblock In \emph{ECCV}, 2018.

\bibitem[Schlichtkrull et~al.(2018)Schlichtkrull, Kipf, Bloem, {van den Berg},
  Titov, and Welling]{Schlichtkrull/etal/2018}
M.~S. Schlichtkrull, T.~N. Kipf, P.~Bloem, R.~{van den Berg}, I.~Titov, and
  M.~Welling.
\newblock Modeling relational data with graph convolutional networks.
\newblock In \emph{ESWC}, 2018.

\bibitem[Sen et~al.(2008)Sen, Namata, Bilgic, and Getoor]{Sen/etal/2008}
G.~Sen, G.~Namata, M.~Bilgic, and L.~Getoor.
\newblock Collective classification in network data.
\newblock \emph{AI Magazine}, 29, 2008.

\bibitem[Shchur et~al.(2018)Shchur, Mumme, Bojchevski, and
  G{\"u}nnemann]{Shchur/etal/2018}
O.~Shchur, M.~Mumme, A.~Bojchevski, and S.~G{\"u}nnemann.
\newblock Pitfalls of graph neural network evaluation.
\newblock In \emph{NeurIPS-W}, 2018.

\bibitem[Simonovsky \& Komodakis(2017)Simonovsky and
  Komodakis]{Simonovsky/Komodakis/2017}
M.~Simonovsky and N.~Komodakis.
\newblock Dynamic edge-conditioned filters in convolutional neural networks on
  graphs.
\newblock In \emph{CVPR}, 2017.

\bibitem[Thekumparampil et~al.(2018)Thekumparampil, Wang, Oh, and
  Li]{Thekumparampil/etal/2018}
K.~K. Thekumparampil, C.~Wang, S.~Oh, and L.~Li.
\newblock Attention-based graph neural network for semi-supervised learning.
\newblock \emph{CoRR}, abs/1803.03735, 2018.

\bibitem[Veli{\v{c}}kovi{\'{c}} et~al.(2018)Veli{\v{c}}kovi{\'{c}}, Cucurull,
  Casanova, Romero, Li{\`{o}}, and Bengio]{Velickovic/etal/2018}
P.~Veli{\v{c}}kovi{\'{c}}, G.~Cucurull, A.~Casanova, A.~Romero, P.~Li{\`{o}},
  and Y.~Bengio.
\newblock Graph attention networks.
\newblock In \emph{ICLR}, 2018.

\bibitem[Veli{\v{c}}kovi{\'{c}} et~al.(2019)Veli{\v{c}}kovi{\'{c}}, Fedus,
  Hamilton, Li{\`{o}}, Bengio, and Hjeml]{Velickovic/etal/2019}
P.~Veli{\v{c}}kovi{\'{c}}, W.~Fedus, W.~L. Hamilton, P.~Li{\`{o}}, Y.~Bengio,
  and R.~D. Hjeml.
\newblock Deep graph infomax.
\newblock In \emph{ICLR}, 2019.

\bibitem[Vinyals et~al.(2016)Vinyals, Bengio, and Kudlur]{Vinyals/etal/2016}
O.~Vinyals, S.~Bengio, and M.~Kudlur.
\newblock Order matters: Sequence to sequence for sets.
\newblock In \emph{ICLR}, 2016.

\bibitem[Wang et~al.(2018{\natexlab{a}})Wang, Yu, Gan, Zheng, Gai, Ye, Li,
  Zhou, Huang, Zhao, Lin, Ma, Deng, Guo, Zhang, Li, Smola, and
  Zhang]{Wang/etal/2018b}
M.~Wang, L.~Yu, A.~Gan, D.~Zheng, Y.~Gai, Z.~Ye, M.~Li, J.~Zhou, Q.~Huang,
  J.~Zhao, H.~Lin, C.~Ma, D.~Deng, Q.~Guo, H.~Zhang, J.~Li, A.~J. Smola, and
  Z.~Zhang.
\newblock Deep graph library.
\newblock \url{http://dgl.ai}, 2018{\natexlab{a}}.

\bibitem[Wang et~al.(2018{\natexlab{b}})Wang, Sun, Liu, Sarma, Bronstein, and
  Solomon]{Wang/etal/2018a}
Y.~Wang, Y.~Sun, Z.~Liu, S.~E. Sarma, M.~M. Bronstein, and J.~M. Solomon.
\newblock Dynamic graph {CNN} for learning on point clouds.
\newblock \emph{CoRR}, abs/1801.07829, 2018{\natexlab{b}}.

\bibitem[Wu et~al.(2019)Wu, Zhang, de~Souza~Jr., Fifty, Yu, and
  Weinberger]{Wu/etal/2019}
F.~Wu, T.~Zhang, A.~H. de~Souza~Jr., C.~Fifty, T.~Yu, and K.~Q. Weinberger.
\newblock Simplifying graph convolutional networks.
\newblock \emph{CoRR}, abs/1902.07153, 2019.

\bibitem[Wu et~al.(2015)Wu, Song, Khosla, Yu, Zhang, Tang, and
  Xiao]{Wu/etal/2015}
Z.~Wu, S.~Song, A.~Khosla, F.~Yu, L.~Zhang, X.~Tang, and J.~Xiao.
\newblock {3D} {ShapeNets}: A deep representation for volumetric shapes.
\newblock In \emph{CVPR}, 2015.

\bibitem[Xu et~al.(2018)Xu, Li, Tian, Sonobe, Kawarabayashi, and
  Jegelka]{Xu/etal/2018}
K.~Xu, C.~Li, Y.~Tian, T.~Sonobe, K.~Kawarabayashi, and S.~Jegelka.
\newblock Representation learning on graphs with jumping knowledge networks.
\newblock In \emph{ICML}, 2018.

\bibitem[Xu et~al.(2019)Xu, Hu, Leskovec, and Jegelka]{Xu/etal/2019}
K.~Xu, W.~Hu, J.~Leskovec, and S.~Jegelka.
\newblock How powerful are graph neural networks?
\newblock In \emph{ICLR}, 2019.

\bibitem[Ying et~al.(2018)Ying, You, Morris, Ren, Hamilton, and
  Leskovec]{Ying/etal/2018}
R.~Ying, J.~You, C.~Morris, X.~Ren, W.~Hamilton, and J.~Leskovec.
\newblock Hierarchical graph representation learning with differentiable
  pooling.
\newblock In \emph{NeurIPS}, 2018.

\bibitem[Zhang et~al.(2018)Zhang, Cui, Neumann, and Chen]{Zhang/etal/2018}
M.~Zhang, Z.~Cui, M.~Neumann, and Y.~Chen.
\newblock An end-to-end deep learning architecture for graph classification.
\newblock In \emph{AAAI}, 2018.

\end{thebibliography}
\bibliographystyle{iclr2019_conference}

\newpage

\begin{appendices}

  \section{Gather and Scatter Operations}%
  \label{sec:gather_and_scatter}

  \begin{figure}[t]
    \centering
    \includegraphics[width=\linewidth]{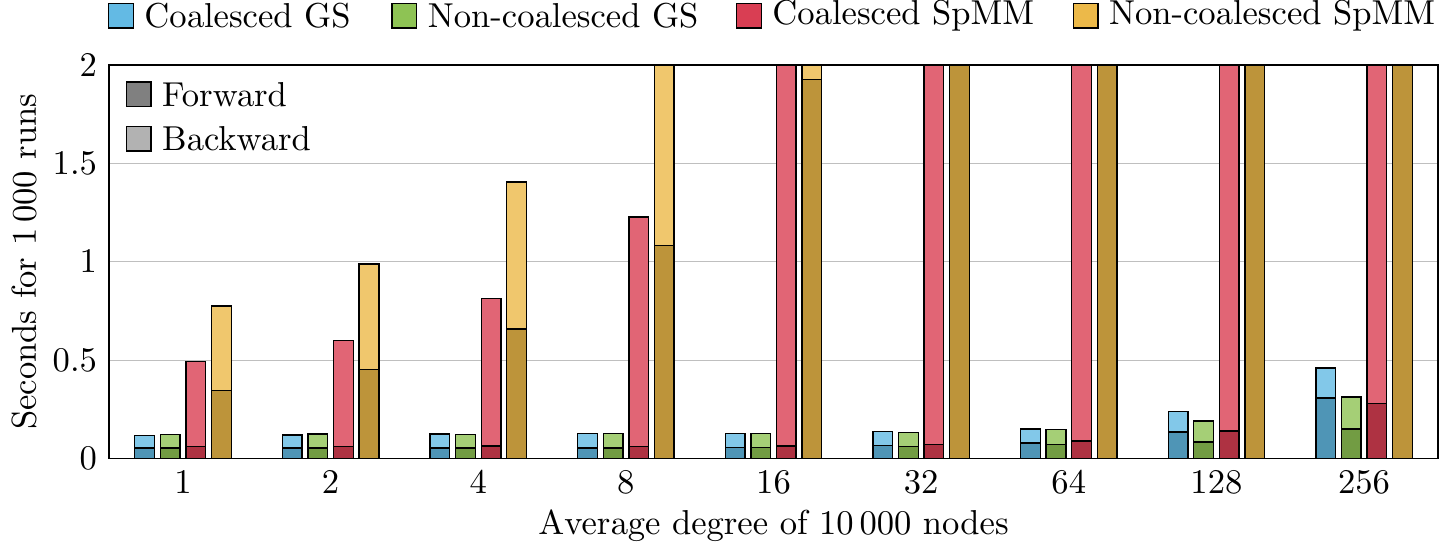}
    \caption{Forward and backward runtimes of $1\,000$ runs of gather and scatter operations (GS) in comparison to sparse-matrix multiplications (SpMM) on Erd\H{o}s Rényi graphs with $10\,000$ nodes and varying average node degrees. Runtimes are capped at two seconds due to visualization. We report runtimes for both coalesced (\ie~ordered by row index) and non-coalesced sparse layout.}\label{fig:bargraph}
  \end{figure}

  PyTorch Geometric makes heavy usage of gather and scatter operations to map node and edge information into edge and node parallel space, respectively.
  Despite inevitable non-coalesced memory access, these operations achieve high data throughput by leveraging parallelization over \emph{all} elements and making use of atomic operations with approximately constant runtime in practice.
  Following upon the PyTorch \texttt{scatter\_add} implementation, we provide our own \texttt{mean} and \texttt{max} operations to allow for all different kinds of aggregation.\footnote{GitHub repository: \url{https://github.com/rusty1s/pytorch\_scatter}}

  Figure~\ref{fig:bargraph} compares the runtime of gather and scatter operations (GS) to the frequently used alternative of using sparse-matrix multiplications (SpMM).
  It shows that atomic operations only begin to throttle the runtime when scattering graphs with high average node degree ($\approx 128$) and even exceed the runtime of highly optimized SpMM executions, both for forward and backward passes.

  Due to SpMM first converting adjacency matrices into \emph{Compressed Row Storage} (CSR) format, it expects \emph{coalesced} sparse tensors (\ie~ordered by row index) which is expensive to compute on GPUs and should be hence performed as part of the pre-processing.
  However, for the backward pass of SpMM, coalescing is performed in any case due to the need of transposing the sparse tensor.
  In contrast, GS is always fast, nevertheless of the input being coalesced.
  Additionally, it allows for modifications of the graph connectivity (\ie~adding self-loops), allows bidirectional data flow, and does naturally support the integration of central node and multi-dimensional edge information.

  However, we do think that our GS scheme can still be improved, \eg, in highly dense graph settings and towards reducing the memory footprint in the edge parallel space.
  In addition, it should be noted that scatter operations are non-deterministic by nature on the GPU\@.
  Although we did not observe any deviations for inference, training results can vary across the same manual seeds.

  \section{Datasets}%
  \label{sec:datasets}

  \begin{table}[t]
    \centering
    \caption{Statistics of the datasets used in the experiments.}\label{tab:datasets}
    \begin{tabular}{lrrrrrrr}
      \toprule
        \textbf{Dataset} & \textbf{Graphs} & \textbf{Nodes} & \textbf{Edges} & \textbf{Features} & \textbf{Classes} & \textbf{Label rate} \\
      \midrule
        Cora     & 1 & 2,708  & 5,278  & 1,433 & 7 & 0.052 \\
        CiteSeer & 1 & 3,327  & 4,552  & 3,703 & 6 & 0.036 \\
        PubMed   & 1 & 19,717 & 44,324 & 500   & 3 & 0.003 \\
      \midrule
        MUTAG    & 188   & 17.93  & 19.79    & 7  & 2 & 0.800 \\
        PROTEINS & 1,113 & 39.06  & 72.82    & 3  & 2 & 0.800 \\
        COLLAB   & 5,000 & 74.49  & 2,457.22 & --- & 3 & 0.800 \\
        IMDB-BINARY   & 1,000 & 19.77  & 96.53    & --- & 2 & 0.800 \\
        REDDIT-BINARY & 2,00  & 429.63 & 497.754  & --- & 2 & 0.800 \\
      \midrule
        ModelNet10 & 4,899 & 1,024 & $\sim$19,440 & --- & 10 & 0.815 \\
      \bottomrule
    \end{tabular}
  \end{table}

  We give detailed descriptions and statistics (\cf~Table~\ref{tab:datasets}) of the datasets used in our experiments:

  \paragraph{Citation Networks.}%

  In the citation network datasets Cora, Citeseer and Pubmed nodes represent documents and edges represent citation links.
  The networks contain bag-of-words feature vectors for each document.
  We treat the citation links as (undirected) edges.
  For training, we use 20 labels per class.

  \paragraph{Social Network Datasets.}%

  COLLAB is derived from three public scientific collaboration datasets.
  Each graph corresponds to an ego-network of different researchers from each field with the task to label each graph to the field the corresponding researcher belongs to.
  IMDB-BINARY is a movie collaboration dataset where each graph corresponds to an ego-network of actors/actresses.
  An edge is drawn between two actors/actresses if they appear in the same movie.
  The task is to infer the genre of the graph.
  REDDIT-BINARY is an online discussion dataset where each graph corresponds to a thread.
  An edge is drawn between two users if one of them responded to another's comment.
  The task is to label each graph to the community/subreddit it belongs to.

  \paragraph{Bioinformatic Datasets.}%

  MUTAG is a dataset consisting of mutagenetic aromatic and heteroaromatic nitro compounds.
  PROTEINS holds a set of proteins represented by graphs.
  Nodes represent secondary structure elements (SSEs) which are connected whenever there are neighbors either in the amino acid sequence or in 3D space.

  \paragraph{3D Object Datasets.}%

  ModelNet10 is an orientation-aligned dataset of CAD models.
  Each model corresponds to exactly one out of 10 object categories.
  Categories were chosen based on a list of the most common object categories in the world.

\end{appendices}

\end{document}